\documentclass[letterpaper,10pt,journal]{IEEEtran}

\usepackage{graphicx}
\usepackage{booktabs}
\usepackage{multirow}
\usepackage{subcaption}
\usepackage{amsmath,amssymb}
\usepackage{cite}
\usepackage{url}
\usepackage{algorithm}
\usepackage{algorithmic}
\usepackage{hyperref}
\usepackage{enumitem}
\setlist[itemize]{leftmargin=*,nosep}

\newcommand{\method}{AULLM++}

\begin{document}

\title{\method: Structured-Token-Conditioned Large Language Models for Micro-Expression Action Unit Detection}

\author{Zhishu Liu, Kaishen Yuan, Bo Zhao, Hui Ma, and Zitong Yu, \IEEEmembership{Senior Member, IEEE}
\thanks{Zhishu Liu and Kaishen Yuan contributed equally to this work.}
\thanks{Corresponding authors: Zitong Yu and Hui Ma.}
\thanks{Z. Liu, B. Zhao, H. Ma, and Z. Yu are with Great Bay University, Dongguan, China.}
\thanks{H. Ma is also with Shenzhen International Graduate School, Tsinghua University, Shenzhen, China.}
\thanks{K. Yuan is with The Hong Kong University of Science and Technology (Guangzhou), Guangzhou, China.}%
}
\maketitle

\begin{abstract}Micro-expression action unit (AU) detection aims to identify localized AUs from subtle, transient facial muscle activations, providing a foundation for decoding detailed affective cues. Existing methods face three key limitations: (1) heavy reliance on weak visual evidence under extremely low signal-to-noise ratios, which makes discriminative cues vulnerable to background noise; (2) coarse-grained feature processing, which is misaligned with the task's demand for fine-grained representations; and (3) limited modeling of inter-AU correlations, which restricts the ability to parse complex expression patterns. In this paper, we propose AULLM++, a structured-token-conditioned LLM framework for micro-expression AU detection. Instead of relying on explicit natural-language reasoning, AULLM++ injects compact visual evidence tokens and AU-relation instruction tokens into a frozen LLM backbone, enabling AU-relation-guided implicit inference for multi-label prediction. The framework consists of three stages: visual evidence construction, AU structure modeling, and token-conditioned AU prediction. Specifically, in the visual branch, we design a Multi-Granularity Evidence-Enhanced Fusion Projector (MGE-EFP) to fuse subtle mid-level texture cues with high-level semantic information, distilling them into a compact Content Token (CT). Inspired by the subset correspondence between micro- and macro-expression AU combinations, we further encode AU relationships as a sparse structural prior and learn instance-conditioned interaction strengths via a Relation-Aware Action Unit Graph Neural Network (R-AUGNN), thereby producing an Instruction Token (IT). In the prompt branch, we fuse CT and IT into a structured textual prompt and introduce Counterfactual Consistency Regularization (CCR) to construct diverse counterfactual variants, thereby improving the model's generalization capability. Extensive experiments demonstrate that AULLM++ achieves competitive performance on standard benchmarks and improves generalization in cross-domain evaluations. The code will be made available at \url{https://github.com/ZS-liu-JLU/AULLMplusplus}.
 
\end{abstract}


\section{Introduction}

\begin{figure}[t]
\centering
\includegraphics[width=\columnwidth]{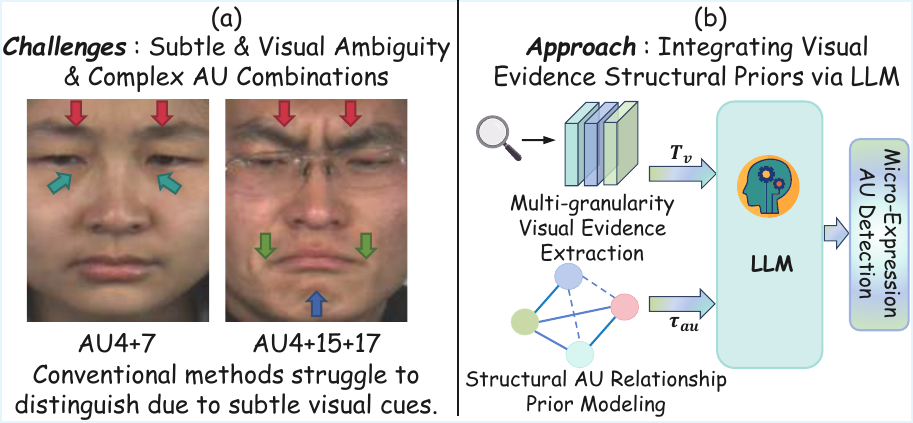}
\caption{
(a) Micro-expression AU detection is challenged by subtle intensity and visually similar AU mixtures (e.g., AU4+7 vs. AU4+15+17), which leads to ambiguity for conventional detectors.
(b) We integrate multi-granularity visual evidence with a structured AU-relation prior and use LLM-based structured-token inference to produce AU predictions, yielding improved cross-dataset generalization.}
\label{fig:teaser}
\vspace{-20pt}
\end{figure}

\IEEEPARstart{D}{efined} by the Facial Action Coding System (FACS)\cite{FACS}, facial action units directly reflect the activation state of local facial muscles and serve as a fundamental basis for affective computing and behavioral analysis. While action unit recognition for macro-expressions has achieved substantial progress, detecting AUs in micro-expressions remains a formidable challenge due to their involuntary nature, short duration, and extremely low intensity\cite{auformer,AAFAA}. The visual evidence of micro-expressions typically appears as fleeting local texture or boundary perturbations around facial muscles, resulting in an extremely low signal-to-noise ratio at the computational level. Under such weak signal amplitudes, discriminative features are susceptible to being overshadowed by background noise such as subject identity variations, illumination changes, and subtle head movements\cite{opticalflow}. As illustrated in Fig.~\ref{fig:teaser}(a), this subtlety often leads to severe visual ambiguity, making it difficult for conventional detectors to distinguish between complex, visually similar AU combinations (e.g., AU4+7 versus AU4+15+17). Therefore, accurately capturing these fundamental muscle movements from barely perceptible visual perturbations remains a core problem in micro-expression analysis.

To address the challenges of feature extraction in micro-expressions, existing research predominantly relies on hand-designed spatio-temporal operators or conventional end-to-end 3D convolutional networks for multi-label feature mining\cite{lbp,3dcnn,3dcnnau}. However, under the subtle visual conditions of micro-expressions, these standard paradigms show clear limitations in feature representation and relation modeling. Current methods often over-rely on sparse visual evidence when learning from features with extremely low signal-to-noise ratios. Whether calculating global optical flow fields or extracting coarse-grained pooled features through deep networks, such processing strategies are misaligned with the intrinsic need for fine-grained representations in micro-expression analysis\cite{opticalflow,opticalmer}. The instantaneous activation of subtle muscles depends heavily on precise high-frequency local evidence, whereas conventional mechanisms can smooth out these important subtle cues. Consequently, the extracted discriminative evidence becomes fragile and is easily dominated by background noise.

At the same time, the anatomical structure of facial muscles determines the natural synergistic or mutually inhibitory relationships between action units\cite{roiau,li2019semantic}. Many existing approaches still treat action unit detection as a set of mutually independent classification tasks, severing this inherent physical correlation and limiting the model's ability to parse complex and rare expression combinations. Although recent studies have introduced graph neural networks to model AU co-occurrence dependencies, they still face bottlenecks in this setting\cite{xie2020assisted}. Because micro-expression datasets are limited in scale and exhibit severe long-tail distributions, traditional graph models often face a trade-off: they either rely on fixed topologies that fail to generalize across unknown subjects and acquisition environments, or perform fully data-driven topology learning on small samples and thus easily overfit dataset-specific biases. In other words, existing relation modeling generally lacks a flexible mechanism that can integrate anatomical priors while adapting to the visual features of the current instance.

Given the extremely low signal-to-noise ratio and complex co-occurrence distribution of micro-expression visual signals, relying solely on data-driven visual pattern matching is insufficient for robust multi-label classification. Therefore, we reformulate micro-expression AU detection from a purely visual feature classification problem into a structured-token-conditioned prediction problem that jointly uses visual evidence and AU-relation priors. Large language models (LLMs) provide powerful pre-trained Transformer backbones for modeling structured token sequences\cite{llmsurvey,blip2}. By mapping subtle visual features and facial anatomical priors into compact continuous tokens, the frozen LLM can be used as a structured inference backbone for AU-relation-guided representation modeling. As depicted in Fig.~\ref{fig:teaser}(b), this paper proposes a structured-token-conditioned micro-expression action unit detection framework named AULLM++, which decomposes the classification task into three stages: evidence construction, structure modeling, and token-conditioned AU prediction.

Specifically, during the visual evidence construction stage, we design a Multi-Granularity Evidence-Enhanced Fusion Projector (MGE-EFP) to capture discriminative cues from subtle muscle movements. This module integrates mid-level features representing local high-frequency texture changes with high-level semantic features carrying global context, compressing them into compact visual content tokens. In the structure modeling stage, inspired by the subset correspondence between macro- and micro-expression action unit combinations, we propose a Relation-Aware Action Unit Graph Neural Network (R-AUGNN). This component injects FACS anatomical priors as a sparse topological structure and adaptively learns interaction weights among action units based on the current input instance, thereby generating AU-relation instruction tokens. During the structured inference stage, the frozen LLM receives a structured sequence composed of the task prompt, content token, and AU-relation instruction tokens. The final hidden representation is then used by a lightweight classification head to produce multi-label AU predictions. Furthermore, to mitigate overfitting to dataset-specific statistical biases, we introduce Counterfactual Consistency Regularization (CCR) during training\cite{cf}. By applying targeted structural perturbations to the AU-relation instruction tokens and constraining non-target AU predictions to remain stable, this regularization strategy enhances the generalization ability and robustness of the network in cross-domain scenarios.

In summary, our main contributions are fourfold:
\begin{itemize}
\item We propose a structured-token-conditioned LLM framework for micro-expression action unit detection. This architecture moves beyond pure visual feature pooling and multi-label classification by introducing a frozen LLM as a structured inference backbone, thereby combining visual evidence with AU-relation priors for robust multi-label AU prediction.
\item We design the multi-granularity evidence-enhanced fusion projector (MGE-EFP) to address the extremely low signal-to-noise ratio of micro-expressions. This component merges mid-level representations of local high-frequency textures with high-level global semantics to extract compact and semantically aligned visual content tokens, reducing the dilution of subtle signals.
\item We develop a relation-aware action unit graph neural network (R-AUGNN). By injecting anatomical priors as a sparse topology and incorporating an instance-adaptive graph learning mechanism, this network explicitly models the synergistic and mutually inhibitory relationships among action units to generate AU-relation instruction tokens for structured-token-conditioned prediction.
\item We introduce counterfactual consistency regularization (CCR) alongside comprehensive benchmarking. Operating exclusively during training, this regularization strategy mitigates overfitting caused by data distribution shifts through targeted perturbations on AU-relation instruction tokens. Extensive experiments demonstrate that AULLM++ achieves the best Macro-F1 among the compared methods in both leave-one-subject-out cross-validation and cross-domain evaluations across three major datasets.
\end{itemize}

\vspace{1ex}
\noindent\textbf{Extensions from Conference Version.} This paper is a substantially extended version of our prior preliminary conference publication, AULLM~\cite{aullm_conf}. While the conference version pioneered the integration of LLMs for micro-expression AU detection by projecting concatenated visual features into the language model space, it treated different AUs as isolated classification targets and lacked explicit mechanisms for handling severe cross-domain distribution shifts. To address these limitations, this manuscript advances the preliminary work into a more robust framework, AULLM++, with the following major improvements:

1. \textbf{Structural Priors:} We introduce R-AUGNN to inject FACS anatomical rules, shifting the paradigm from isolated recognition to relation-aware structured inference.
2. \textbf{Visual Enhancement:} We upgrade the visual frontend to MGE-EFP, explicitly decoupling high-frequency muscle motion from global semantics to form more precise visual tokens.
3. \textbf{Intervention-inspired Robustness:} We propose CCR during training to mitigate dataset-specific shortcut correlations, thereby improving cross-domain generalization.
4. \textbf{Comprehensive Evaluation:} We conduct more extensive experiments, including rigorous cross-domain evaluations and new validation on the challenging 4DME-Micro dataset\cite{fourdme}.

\section{Related Work}
\label{sec:related_work}
\subsection{Micro-Expression Action Unit Detection}
Early micro-expression analysis heavily relied on hand-designed spatio-temporal feature operators. For instance, classical methods such as LBP-TOP~\cite{lbp}, histograms of oriented optical flow (HOOF), and optical strain were widely used to capture subtle facial dynamics. With the rapid development of deep learning in recent years, spatio-temporal convolutional networks and Vision Transformers (e.g., AUFormer~\cite{auformer}) have gradually become mainstream, driving feature extraction toward an end-to-end learning paradigm.

To tackle the specific challenge of weak micro-expression features, researchers have designed various targeted mechanisms. For example, Li et al.~\cite{sca} introduced a Spatial and Channel Attention (SCA) network to focus on critical facial regions, and further proposed a Dual-View Attentive Similarity-Preserving (DVASP)~\cite{dvasp} knowledge distillation strategy to enhance the discriminative power of action units at the feature level. Eulerian video magnification techniques have also been incorporated to amplify visually imperceptible movements. Recent representative works, such as the Learnable Eulerian Dynamics (LED) module proposed by Varanka et al.~\cite{LED}, combine frequency-domain filters with deep networks to adaptively enhance subtle motions. To address the background and illumination noise introduced by magnification, Khor et al.~\cite{infusenet} developed infused suppression mechanisms to purify authentic weak motion signals from magnification artifacts. More recently, Zhou et al.~\cite{zhou2025objective} explored objective class-based micro-expression recognition through simultaneous AU detection and feature aggregation, highlighting the benefit of incorporating AU-level information into micro-expression analysis. Wei et al.~\cite{wei2026micro} further proposed Micro-AU CLIP, which introduces fine-grained contrastive learning from local independence to global dependency for micro-expression AU detection. These recent studies indicate that AU-level structural modeling remains an active and important direction for fine-grained micro-expression understanding.

Despite these attempts to improve the signal-to-noise ratio, existing methods still exhibit considerable limitations under extremely low-intensity signals. Conventional dense convolutions or global pooling operations can conflate local high-frequency textures with global background noise, leading to the loss of fine-grained information\cite{resnet,sca}. Moreover, feeding high-dimensional dense visual features directly into downstream modules incurs high computational costs and can dilute weak discriminative signals during complex forward propagation\cite{blip2,zhu2025uniemo}. Motivated by this, we design the Multi-Granularity Evidence-Enhanced Fusion Projector (MGE-EFP). By decoupling and re-fusing mid-level features representing local high-frequency textures and high-level features depicting global semantics, this module compresses multi-dimensional spatio-temporal evidence into compact and semantically aligned content tokens. This design filters redundant noise while retaining fine-grained discriminative cues for subsequent structured-token-conditioned inference.

\subsection{Facial Relation Modeling and Physical Priors}
Facial action units are driven by the anatomical structure of facial skeletal muscles and exhibit strong combinational patterns. Early research mostly treated AU detection as a series of independent binary classification tasks or relied on simple multi-label regression (e.g., Zhao et al.~\cite{au_cnn}), often ignoring the inherent physiological dependencies among action units. Recently, graph neural networks (GNNs) have been widely applied to capture these topological relationships. In the micro-expression domain, representative works have also emerged\cite{megacov}. For instance, Xie et al.~\cite{xie2020assisted} proposed an AU-assisted graph attention convolutional network to model spatial and temporal interactions between facial regions for micro-expression recognition. Similarly, Li et al.~\cite{li2019semantic} used semantic relationships to guide representation learning for AU recognition.

Despite these advances, modeling graph relations specifically for micro-expression AU detection remains challenging. Because micro-expression datasets are limited in scale and exhibit severe long-tail class distributions, traditional graph models often face a difficult trade-off. Fixed static topologies can be too rigid to adapt to individual differences and dynamic variations among subjects, whereas fully data-driven dynamic graph learning mechanisms are prone to overfitting the statistical biases of small-sample datasets.

To address this bottleneck, our proposed Relation-Aware Action Unit Graph Neural Network (R-AUGNN) directly targets the detection task itself. Rather than relying on purely data-driven topologies, this module injects Facial Action Coding System (FACS) anatomical rules as a sparse prior topology. It then adaptively infers interaction weights based on the micro-visual features of the current instance, generating AU-relation instruction tokens that provide structured prior guidance for the subsequent LLM-based inference module.

\subsection{LLM-based Structured Inference and Robust Representation Learning}
With the rapid development of vision-language models (e.g., BLIP-2~\cite{blip2} and LLaVA~\cite{llava}), Large Language Models (LLMs) have demonstrated strong potential in multimodal representation learning. In affective computing, recent approaches typically convert visual features into soft prompts for LLMs to output emotion labels. However, micro-expression detection is a fine-grained task that relies on subtle visual evidence. Standard visual reasoning paradigms---such as feeding large numbers of image patches directly into an LLM---may marginalize already weak micro-expression signals during long-context attention calculations~\cite{zhu2025emosym,expllmfacial,zhu2026H-GAR}. Consequently, AULLM++ constructs an efficient tokenized interface. By inputting compressed content tokens and structural instruction tokens into the LLM, we use the frozen LLM as a structured-token-conditioned inference backbone, enabling AU-relation-guided representation modeling instead of explicit natural-language reasoning.

\begin{figure*}[t]
\centering
\includegraphics[width=\textwidth]{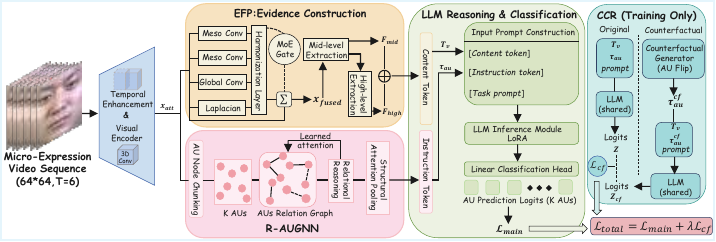}
\caption{Overall architecture of AULLM++. The model constructs a compact visual evidence token $T_v$ and a FACS-guided AU-structure instruction token $\tau_{au}$, injects them into an LLM backbone for structured-token-conditioned AU classification, and applies CCR during training for robustness.}
\label{fig:framework}
\end{figure*}

Furthermore, real-world micro-expression recognition suffers from severe domain shifts. Traditional countermeasures mostly focus on domain adaptation (e.g., adversarial training such as DANN~\cite{dann}), which implicitly aligns feature distributions but may still retain dataset-specific shortcut correlations. Recently, counterfactual learning has emerged as a useful tool for reducing spurious correlations in computer vision~\cite{cf}. However, performing counterfactual interventions directly in pixel space can destroy the integrity of the original micro-expression signals. To address this issue, our Counterfactual Consistency Regularization (CCR) applies targeted structural perturbations at the AU-relation instruction level during training, encouraging AU-specific structural sensitivity while maintaining stable predictions for non-target AUs. In this work, CCR is regarded as an intervention-inspired regularization strategy rather than a strict physical counterfactual generation module, and it introduces no additional overhead during inference.

Finally, our preliminary conference version, AULLM~\cite{aullm_conf}, made an early attempt to apply LLM-based token-conditioned inference to micro-expression detection. Despite this early contribution, AULLM relied on concatenated visual features without explicit AU-relation constraints, making it vulnerable to domain shifts. The proposed AULLM++ advances this framework by integrating relation-aware graph modeling (R-AUGNN) and intervention-inspired consistency regularization (CCR). In contrast to fragmented pipelines, AULLM++ constructs a unified system that maps subtle facial dynamics and physical priors into structured visual and relational tokens, addressing the technical bottlenecks of weak features, rigid relations, and cross-domain fragility.

\section{Methodology}
\label{sec:methodology}

This section presents the theoretical foundation and technical details of the proposed AULLM++ framework. We first provide a mathematical formulation of the micro-expression action unit detection task and an overview of the architecture. We then describe the computational mechanisms of the visual evidence construction module and the structural relation graph \cite{gnn}. Finally, we introduce the LLM-based structured inference process and the optimization objective of Counterfactual Consistency Regularization (CCR)\cite{cfreview}.

\vspace{1.5ex}
\noindent\textbf{Problem Formulation and Overall Architecture.} Micro-expression action unit detection is fundamentally a multi-label binary classification problem under an extremely low signal-to-noise ratio. Given a micro-expression dataset $\mathcal{D} = \{(\mathbf{X}^{(i)}, \mathbf{y}^{(i)})\}_{i=1}^M$ containing $M$ samples, the input for the $i$-th instance is a face-aligned and cropped micro-expression video clip $\mathbf{X} \in \mathbb{R}^{T \times C \times H \times W}$, where $T$ denotes the sequence length, $C$ is the number of channels, and $H$ and $W$ represent the spatial resolution. The target label is defined as $\mathbf{y} = [y_1, y_2, \dots, y_N]^\top \in \{0, 1\}^N$, where $N$ is the total number of predefined action unit categories. An element $y_k = 1$ indicates that the $k$-th AU is activated in the clip, whereas $0$ denotes its absence. The core objective is to learn a robust non-linear mapping function $\mathcal{F}: \mathbf{X} \rightarrow \hat{\mathbf{y}}$. This function should filter background noise, such as identity variations and illumination fluctuations, while capturing subtle local facial muscle deformations and decoding them into an accurate multi-label probability distribution $\hat{\mathbf{y}} \in [0, 1]^N$.

To address the bottlenecks of traditional pure visual feature mapping under subtle micro-expression signals, the proposed AULLM++ framework reformulates the end-to-end black-box regression paradigm. Instead, it decomposes the mapping function $\mathcal{F}$ into three stages: visual evidence construction, structural instruction modeling, and LLM-based structured inference. The information flow of the entire system is illustrated in Fig.~\ref{fig:framework}. Specifically, in the visual evidence construction stage, the input video sequence is fed into a Multi-Granularity Evidence-Enhanced Fusion Projector (MGE-EFP). This module independently extracts and disentangles mid-level high-frequency features and high-level abstract semantics. After adaptive feature fusion, the system compresses the multi-dimensional spatio-temporal features into a compact visual content token aligned with the hidden dimension of the large model. In parallel, for structural instruction modeling, a Relation-Aware Action Unit Graph Neural Network (R-AUGNN) instantiates predefined AU correlations into a sparse prior topology. Combined with the micro-visual features from the current instance, the network adaptively infers an instruction token containing dynamic interaction weights. Finally, during the structured inference and regularization stage, the visual content token and the structural instruction token are embedded into a structured input prompt. The frozen large language model receives this joint prompt and models the structured token sequence for AU prediction. During training, CCR applies targeted structural perturbations to encourage AU-specific structural sensitivity and enhance cross-domain robustness.

\vspace{1.5ex}
\noindent\textbf{Multi-Granularity Evidence-Enhanced Fusion Projector.} The central challenge of micro-expression visual signals is that subtle muscle boundary deformations are typically concealed within local high-frequency textures, whereas understanding the overall facial posture requires global low-frequency abstract features\cite{se,liu2025temporal}. Conventional single-scale pooling operations can cause high-frequency micro-cues to be overshadowed by global semantics\cite{SHAO}. To address this issue, MGE-EFP extracts purified visual evidence by disentangling and re-fusing spatio-temporal information under different receptive fields.

Given the input sequence $\mathbf{X}$, we first perform hierarchical feature encoding with a lightweight 3D spatio-temporal backbone network. Let the mid-level spatio-temporal tensor extracted by the network be $\mathbf{F}_{mid} \in \mathbb{R}^{T' \times C_1 \times H' \times W'}$, and the high-level semantic tensor be $\mathbf{F}_{high} \in \mathbb{R}^{T' \times C_2 \times H'' \times W''}$. To amplify instantaneous facial texture perturbations when micro-expressions occur, we introduce a differentiable Laplacian enhancement operator $\nabla^2$ along the spatial dimensions to apply high-frequency excitation to the mid-level features. The high-frequency micro-feature representation $\mathbf{F}_{hf}$ is obtained as follows:
\begin{equation}
\mathbf{F}_{hf} = \mathbf{F}_{mid} + \gamma \cdot \nabla^2(\mathbf{F}_{mid}),
\end{equation}
where $\gamma$ is a learnable adaptive enhancement coefficient that controls the injection intensity of high-frequency textures. This operation magnifies minute gradient changes along the edges of muscle movements.

To integrate global semantic context while preserving high-frequency details, we introduce an attention-gated mechanism. Spatial downsampling alignment is first applied to $\mathbf{F}_{hf}$ to match the spatio-temporal resolution of $\mathbf{F}_{high}$, yielding $\tilde{\mathbf{F}}_{hf}$. We then generate an adaptive fusion weight matrix $\mathbf{G} \in (0, 1)$ via a cross-channel gating function:
\begin{equation}
\mathbf{G} = \sigma \left( \mathbf{W}_{g1} \ast \tilde{\mathbf{F}}_{hf} + \mathbf{W}_{g2} \ast \mathbf{F}_{high} + \mathbf{b}_{g} \right),
\end{equation}
where $\sigma(\cdot)$ is the sigmoid activation function, and $\mathbf{W}_{g1}, \mathbf{W}_{g2}$ are convolutional kernels. This gating matrix determines the fusion ratio between local micro-textures and global semantic features at the pixel level. The fused multi-granularity feature $\mathbf{F}_{fused}$ is formulated as:
\begin{equation}
\mathbf{F}_{fused} = \mathbf{G} \odot \tilde{\mathbf{F}}_{hf} + (1 - \mathbf{G}) \odot \mathbf{F}_{high},
\end{equation}
where $\odot$ denotes the Hadamard product. Finally, to satisfy the structural constraints of the large language model for 1D sequence inputs, we perform spatio-temporal pooling on $\mathbf{F}_{fused}$ and map it to the hidden subspace of the large language model through a non-linear projection matrix $\mathbf{W}_{proj}$. This generates a condensed visual content token $T_v$:
\begin{equation}
T_v = \text{ReLU} \left( \text{Pool}(\mathbf{F}_{fused}) \mathbf{W}_{proj} \right) \in \mathbb{R}^{1 \times D},
\end{equation}
where $D$ is the embedding dimension. This token encapsulates the compact visual evidence after denoising and high-frequency enhancement, serving as the visual basis for subsequent structured-token-conditioned inference.

\vspace{1.5ex}
\noindent\textbf{Relation-Aware Action Unit Graph Neural Network.} Relying solely on visual evidence is prone to misjudgments under the long-tail distribution of micro-expressions\cite{megacov}. The Facial Action Coding System (FACS) indicates that anatomical linkages between muscles provide strong priors for AU activations. R-AUGNN is designed to combine static anatomical priors with dynamic visual instances to construct structured physical instructions.

We define all $N$ target AUs as the node set of a graph. Based on psychological statistics and FACS anatomical rules, we pre-construct a sparse static prior adjacency matrix $\mathbf{A}_{prior} \in \mathbb{R}^{N \times N}$. When a strong synergistic or mutually inhibitory relationship exists between the $i$-th and $j$-th AUs, $\mathbf{A}_{prior}^{(i,j)}$ is assigned an initial connection weight; otherwise, it is set to $0$. Meanwhile, using the high-level features $\mathbf{F}_{high}$ extracted by the visual backbone, we initialize an independent node representation $\mathbf{h}_i^{(0)} \in \mathbb{R}^d$ for each AU through class-specific projections, forming the initial node matrix $\mathbf{H}^{(0)}$.

Considering facial skeletal differences among subjects and the dynamic nature of micro-expressions, a rigid $\mathbf{A}_{prior}$ cannot cover all scenarios. Therefore, we introduce a self-attention mechanism in the node feature space to calculate the dynamic interaction intensity $e_{ij}$ between node $i$ and node $j$ for the current instance:
\begin{equation}
e_{ij} = \text{LeakyReLU} \left( \mathbf{a}^\top [\mathbf{W}_q \mathbf{h}_i^{(0)} \parallel \mathbf{W}_k \mathbf{h}_j^{(0)}] \right),
\end{equation}
where $\parallel$ denotes vector concatenation, and $\mathbf{W}_q, \mathbf{W}_k$, and $\mathbf{a}$ are learnable parameters. By applying softmax normalization to $e_{ij}$, we obtain the instance-adaptive adjacency matrix $\mathbf{A}_{dynamic}$. To balance anatomical rules and instance-specific variation, the system performs a weighted fusion of the static prior and dynamic attention to generate the final relation routing matrix $\hat{\mathbf{A}}$:
\begin{equation}
\hat{\mathbf{A}} = \alpha \mathbf{A}_{prior} + (1 - \alpha) \mathbf{A}_{dynamic},
\end{equation}
where $\alpha$ is a learnable balancing coefficient biased toward the FACS prior.

With the relation graph established, R-AUGNN performs message passing via an $L$-layer Graph Convolutional Network. The information update for the $l$-th layer is defined as:
\begin{equation}
\mathbf{H}^{(l+1)} = \sigma \left( \tilde{\mathbf{D}}^{-\frac{1}{2}} \hat{\mathbf{A}} \tilde{\mathbf{D}}^{-\frac{1}{2}} \mathbf{H}^{(l)} \mathbf{W}_{gcn}^{(l)} \right),
\end{equation}
where $\tilde{\mathbf{D}}$ is the degree matrix, and $\mathbf{W}_{gcn}^{(l)}$ is the transition weight matrix. After $L$ iterations of message passing, a Multi-Layer Perceptron (MLP) aligns these relation-aware node features to the semantic space of the large language model, generating the structural instruction token $\tau_{au} = \text{MLP}(\mathbf{H}^{(L)}) \in \mathbb{R}^{N \times D}$. These $N$ structural tokens serve as AU-relation priors for structured-token-conditioned AU prediction.
Since each node is indexed by a predefined AU category and the final multi-label BCE loss supervises the corresponding AU output dimension, the AU-token correspondence is preserved by both construction and classification supervision. Graph propagation enriches each AU token with contextual relation information, but it does not reorder or collapse the AU-indexed node sequence.

\begin{algorithm}[t]
\caption{Training and Inference Pipeline of AULLM++}
\label{alg:pipeline}
\begin{algorithmic}[1]
\REQUIRE Dataset $\mathcal{D}$, FACS prior matrix $\mathbf{A}_{prior}$, LLM prompt prefix $\mathcal{P}_{text}$, pre-trained LLM $\mathcal{M}_{\Phi}$, hyperparameters $\gamma, \alpha, \lambda_{inv}, \lambda_{\Delta}, \lambda_{ccr}$.
\STATE \textbf{// Phase 1: Forward Inference (Training and Testing)}
\FOR{each instance sequence $\mathbf{X}$ in mini-batch}
\STATE Extract features $\mathbf{F}_{mid}, \mathbf{F}_{high}$ via 3D backbone.
\STATE $\mathbf{F}_{hf} \leftarrow \mathbf{F}_{mid} + \gamma \cdot \nabla^2(\mathbf{F}_{mid})$ \COMMENT{High-freq enhancement}
\STATE $\mathbf{F}_{fused} \leftarrow \mathbf{G} \odot \tilde{\mathbf{F}}_{hf} + (1 - \mathbf{G}) \odot \mathbf{F}_{high}$ \COMMENT{Gated fusion}
\STATE $T_v \leftarrow \text{Pool}(\mathbf{F}_{fused}) \mathbf{W}_{proj}$ \COMMENT{Visual content token}
\STATE $\hat{\mathbf{A}} \leftarrow \alpha \mathbf{A}_{prior} + (1 - \alpha) \mathbf{A}_{dynamic}$ \COMMENT{Adaptive routing}
\STATE $\tau_{au} \leftarrow \text{MLP}(\text{GCN}(\mathbf{H}^{(0)}, \hat{\mathbf{A}}))$ \COMMENT{Structural instruction}
\STATE Construct prompt: $\mathbf{E}_{in} \leftarrow [ \mathbf{E}_{text} \parallel T_v \parallel \tau_{au} ]$
\STATE Predict AU probabilities: $\hat{\mathbf{y}} \leftarrow \sigma(\mathcal{M}_{\Phi}(\mathbf{E}_{in}))$
\ENDFOR
\STATE \textbf{// Phase 2: CCR Perturbation (Training Only)}
\STATE Compute base classification loss: $\mathcal{L}_{cls} = \text{BCE}(\hat{\mathbf{y}}, \mathbf{y})$
\FOR{each target AU $k$}
\STATE Apply targeted structural perturbation: $\tilde{\tau}_{au}^{(k)} \leftarrow \tau_{au}^{(k)} + \delta_k$
\STATE Re-forward LLM with $\tilde{\tau}_{au}^{(k)}$ to obtain counterfactual $\hat{\mathbf{y}}^{cf}$.
\STATE Calculate targeted perturbation loss: $\mathcal{L}_{away} = \text{BCE}(\hat{y}_k^{cf}, 1 - y_k)$
\STATE Calculate invariance loss: $\mathcal{L}_{inv} = \sum_{j \neq k} \text{KL}(\hat{y}_j \parallel \hat{y}_j^{cf})$
\STATE $\mathcal{L}_{ccr} \leftarrow \mathcal{L}_{away} + \lambda_{inv} \mathcal{L}_{inv} + \lambda_{\Delta} \|\delta_k\|_2^2$
\ENDFOR
\STATE Update network parameters by minimizing $\mathcal{L}_{total} = \mathcal{L}_{cls} + \lambda_{ccr} \mathcal{L}_{ccr}$.
\end{algorithmic}
\end{algorithm}

\vspace{1.5ex}
\noindent\textbf{LLM-based Structured Inference and Consistency Regularization.} After acquiring the visual content token $T_v$ and the structural instruction token $\tau_{au}$, AULLM++ uses a frozen Large Language Model as a structured-token-conditioned inference backbone for final AU decoding. We first define a task-specific text prompt mapped to sequence features $\mathbf{E}_{text}$. We then concatenate the visual content and structural instructions along the sequence dimension to construct the complete structured input prompt tensor $\mathbf{E}_{in} = [ \mathbf{E}_{text} \parallel T_v \parallel \tau_{au} ]$. To preserve the generalized representation capability while avoiding catastrophic forgetting, we freeze the backbone parameters and inject Low-Rank Adaptation (LoRA) matrices into the attention modules. The LLM receives $\mathbf{E}_{in}$ through input embeddings, and the final hidden state $\mathbf{h}_{out} \in \mathbb{R}^D$ is used as the structured representation for AU classification\cite{visualprompttuning,coop,lora}. AU multi-label prediction probabilities are computed with a linear classification head $\hat{\mathbf{y}} = \sigma ( \mathbf{W}_{cls} \mathbf{h}_{out} + \mathbf{b}_{cls} ) \in [0, 1]^N$.

Micro-expression recognition faces severe distribution shifts in cross-domain environments. Models often fit specific illumination or identity features as statistical shortcuts rather than learning robust AU-related representations. To reduce such shortcut correlations, we introduce an instruction-level Counterfactual Consistency Regularization (CCR) mechanism. As summarized in Algorithm \ref{alg:pipeline}, during training, for a target AU $k$, we apply a targeted perturbation $\delta_k$ to its corresponding structural instruction token, constructing a perturbed instruction $\tilde{\tau}_{au}^{(k)} = \tau_{au}^{(k)} + \delta_k$. The perturbed instruction tokens are re-fed into the LLM to yield a counterfactual prediction $\hat{\mathbf{y}}^{cf}$. The perturbation on the AU-$k$ structural token is not intended to guarantee an exact physical inversion of AU-$k$. Instead, it serves as a targeted sensitivity constraint: the model is encouraged to move away from the original AU-$k$ decision when the AU-$k$-related structural cue is disturbed, while predictions for non-intervened AUs are constrained to remain stable. Thus, the CCR objective $\mathcal{L}_{ccr}$ is defined as:
\begin{equation}
\mathcal{L}_{ccr} = \underbrace{\mathcal{L}_{bce}(\hat{y}_k^{cf}, 1 - y_k)}_{\mathcal{L}_{away}} + \lambda_{inv} \sum_{j \neq k} \text{KL}(\hat{y}_j \parallel \hat{y}_j^{cf}) + \lambda_{\Delta} \|\delta_k\|_2^2,
\end{equation}
where $\text{KL}(\cdot \parallel \cdot)$ is the Kullback-Leibler divergence penalizing minute drifts in non-target distributions, $\|\delta_k\|_2^2$ regulates the perturbation magnitude, and $\lambda_{inv}, \lambda_{\Delta}$ are balancing coefficients.

Throughout the training process, standard Binary Cross-Entropy (BCE) loss serves as the primary classification objective $\mathcal{L}_{cls}$. The global optimization objective of AULLM++ is formulated as:
\begin{equation}
\mathcal{L}_{total} = \mathcal{L}_{cls}(\hat{\mathbf{y}}, \mathbf{y}) + \lambda_{ccr} \mathcal{L}_{ccr}.
\end{equation}
Importantly, CCR acts solely as a regularization constraint during training. During inference, the counterfactual module is removed, ensuring that AULLM++ improves cross-domain robustness without introducing additional inference overhead from CCR.

\section{Experiments}
\label{sec:experiments}
\begin{table*}[t]
  \centering
  \caption{Micro-expression AU detection performance comparison on CASME II under the LOSO protocol. F1-scores (\%) are reported. The best results are highlighted in \textbf{bold}. Avg. denotes Macro-F1.}
  \label{tab:au_casme2_transposed}
  \scriptsize
  \setlength{\tabcolsep}{4pt} 
  \renewcommand{\arraystretch}{1.2} 

  \begin{tabular*}{\textwidth}{@{\extracolsep{\fill}} l ccccccccc @{}}
    \hline\hline
    \textbf{Method} & \textbf{AU1} & \textbf{AU2} & \textbf{AU4} & \textbf{AU7} & \textbf{AU12} & \textbf{AU14} & \textbf{AU15} & \textbf{AU17} & \textbf{Avg.} \\
    \hline
    LBP-TOP~\cite{lbp}      & 80.9 & 60.7 & 89.8 & 56.2 & 69.3 & 63.2 & 52.2 & 75.8 & 68.5 \\
    ResNet18~\cite{resnet}  & 58.9 & 76.2 & 82.7 & 48.7 & 59.3 & 64.2 & 48.1 & 62.9 & 62.6 \\
    ResNet34~\cite{resnet}  & 55.1 & 69.2 & 83.2 & 55.4 & 62.8 & 56.7 & 52.2 & 62.9 & 62.2 \\
    FitNet~\cite{fitnet}    & 65.5 & 74.3 & 79.1 & 46.8 & 61.7 & 62.4 & 53.9 & 57.6 & 62.7 \\
    SP~\cite{sp}            & 69.0 & 68.5 & 77.7 & 51.1 & 57.0 & 65.3 & 56.6 & 61.3 & 63.3 \\
    AT~\cite{at}            & 65.0 & 62.7 & 83.1 & 57.6 & 58.6 & 62.5 & 46.6 & 65.6 & 62.7 \\
    SCA~\cite{sca}          & 64.1 & 76.7 & 81.1 & 56.2 & 61.3 & 61.9 & 68.8 & 64.4 & 66.8 \\
    DVASP~\cite{dvasp}      & 72.6 & 72.1 & 89.8 & 56.9 & 79.6 & 68.5 & 71.5 & 70.0 & 72.6 \\
    \hline
    Res18Net LED~\cite{LED} & 85.8 & 81.7 & 89.0 & 53.9 & 71.4 & 78.5 & 74.5 & \textbf{83.5} & 77.3 \\
    SSSNet LED~\cite{LED}   & 92.6 & 83.7 & 88.6 & 63.7 & 76.6 & 74.4 & 71.5 & 76.1 & 78.4 \\
    \hline
    AULLM (Conf.)~\cite{aullm_conf} & 92.2 & \textbf{88.4} & 87.0 & 68.1 & 79.0 & 75.1 & 79.0 & 81.0 & 81.4 \\
    \textbf{AULLM++ (\mbox{Qwen3.5-0.8B})}& \textbf{93.8} & 85.8 & 88.2 & \textbf{70.7} & 78.7 & \textbf{79.0} & 78.0 & 81.6 & 82.1 \\
    \textbf{AULLM++ (\mbox{Qwen3.5-3B})}& 92.4 & 87.5 & \textbf{91.8} & 62.5 & 78.5 & 76.4 & 75.7 & 83.4 & 81.0 \\
    \textbf{AULLM++ (\mbox{DeepSeek-1.5B})} & 93.1 & 88.1 & 90.5 & 67.1 & \textbf{81.0} & 76.1 & \textbf{80.0} & 82.0 & \textbf{82.4} \\
    \hline\hline
  \end{tabular*}
\end{table*}

To evaluate the effectiveness, structured-token-conditioned inference capability, and cross-domain robustness of the proposed AULLM++ framework, we conduct extensive experiments on three widely used spontaneous micro-expression benchmarks. We systematically compare AULLM++ against representative baselines, including recent motion magnification methods\cite{LED} and our preliminary conference version (AULLM)~\cite{aullm_conf}, and provide ablation studies alongside qualitative analyses.

\subsection{Datasets and Evaluation Protocols}
\textbf{Datasets.} We use three spontaneous micro-expression datasets: CASME II~\cite{yan2014casme}, SAMM~\cite{samm}, and the recently introduced 4DME-Micro~\cite{fourdme}.
CASME II contains 247 spontaneous micro-expression samples elicited from 26 subjects, recorded at 200 frames per second (fps). The samples are rigorously annotated with action units (AUs) based on the Facial Action Coding System (FACS)\cite{FACS}. Following standard practices in the community, we focus on eight main AUs: AU1, AU2, AU4, AU7, AU12, AU14, AU15, and AU17.
SAMM comprises 159 micro-movements from 32 subjects of diverse demographic backgrounds, ensuring a high degree of ethnic diversity.
4DME-Micro is a challenging benchmark that includes spontaneous expressions captured under diverse elicitation paradigms. For all datasets, apex frames and their surrounding temporal sequences are extracted and used as inputs.

\vspace{1.5ex}
\noindent\textbf{Evaluation Protocols.} Micro-expression datasets exhibit skewed long-tail distributions, with certain AUs (e.g., AU12) appearing frequently while others (e.g., AU15) are rare. Consequently, standard accuracy can be misleading and may conceal poor performance on minority classes. To address this issue, we adopt Macro F1-Score (Macro-F1) as the primary evaluation metric. It calculates the unweighted average of F1-scores across all evaluated AUs, providing a stringent measurement of detection capability under imbalanced distributions. Furthermore, micro-expression signals are susceptible to subject-specific physiological traits. As highlighted by recent studies on data leakage and evaluation issues in micro-expression analysis~\cite{varanka2023data}, preventing subject identity overlap is crucial. Therefore, to prevent subject identity leakage and evaluate genuine generalization, all within-domain experiments follow the Leave-One-Subject-Out (LOSO) cross-validation protocol.

\subsection{Implementation Details}
For reproducibility, we detail the implementation configurations of AULLM++.
\begin{table}[t]
  \centering
  \caption{Micro-expression AU detection performance comparison on the SAMM dataset under the LOSO protocol. F1-scores (\%) are reported. The best results are highlighted in \textbf{bold}. Avg. denotes Macro-F1.}
  \label{tab:au_samm_transposed}
  \scriptsize
  \renewcommand{\arraystretch}{1.2} 

  \begin{tabular*}{\columnwidth}{@{\extracolsep{\fill}} l ccccc @{}}
    \hline\hline
    \textbf{Method} & \textbf{AU2} & \textbf{AU4} & \textbf{AU7} & \textbf{AU12} & \textbf{Avg.} \\
    \hline
    LBP-TOP~\cite{lbp}      & 58.8 & 47.9 & 44.5 & 49.5 & 50.2 \\
    ResNet18~\cite{resnet}  & 49.7 & 49.1 & 46.1 & 40.5 & 46.4 \\
    ResNet34~\cite{resnet}  & 44.0 & 55.2 & 38.0 & 40.5 & 44.4 \\
    Fit-18~\cite{fitnet}    & 54.1 & 51.2 & 44.5 & 48.3 & 49.5 \\
    SP-18~\cite{sp}         & 42.8 & 64.2 & 38.1 & 49.5 & 48.7 \\
    AT-18~\cite{at}         & 47.2 & 60.5 & 43.5 & 38.0 & 47.3 \\
    SCA~\cite{sca}          & 45.7 & 59.2 & 43.9 & 53.2 & 50.5 \\
    DVASP-18~\cite{dvasp}   & 47.8 & 67.5 & 48.1 & 44.7 & 52.0 \\
    \hline
    Res18 LED~\cite{LED}    & 57.4 & 71.3 & 53.2 & 47.3 & 57.3 \\
    SSSNet LED~\cite{LED}   & 61.5 & 62.4 & 45.2 & 47.8 & 54.2 \\
    \hline
    AULLM (Conf.)~\cite{aullm_conf} & 66.9 & 71.9 & 55.8 & 52.8 & 61.9 \\
    \textbf{AULLM++ (Ours)} & \textbf{70.6} & \textbf{72.6} & \textbf{56.5} & \textbf{53.5} & \textbf{62.6} \\
    \hline\hline
  \end{tabular*}
\end{table}

\textbf{Visual and LLM Backbone.} For visual evidence extraction, we employ a lightweight 3D Convolutional Neural Network (3D-CNN) backbone\cite{3dcnn}. Given the data scarcity and subtlety of micro-expressions, a lightweight architecture helps reduce the overfitting typically associated with heavily parameterized networks while still extracting spatio-temporal dynamics. For the LLM backbone, we instantiate our framework using \texttt{deepseek-ai/DeepSeek-R1-Distill-Qwen-1.5B}\cite{deepseekr11.5}. This compact open-source LLM supports continuous input embedding injection and LoRA-based adaptation, making it suitable for structured-token-conditioned AU prediction. To align the multimodal tokens while preventing catastrophic forgetting of the LLM's generalized knowledge, we freeze the original backbone parameters and inject Low-Rank Adaptation (LoRA) matrices into the attention modules, configuring the rank as $r=16$ and the scaling factor as $\alpha=32$\cite{lora}. To examine the influence of the LLM backbone, Table~\ref{tab:au_casme2_transposed} also reports CASME II results using \mbox{Qwen3.5-0.8B} and \mbox{Qwen3.5-3B} variants. \mbox{Qwen3.5-0.8B} obtains 82.1\% Macro-F1, \mbox{Qwen3.5-3B} obtains 81.0\%, and \mbox{DeepSeek-1.5B} obtains the best average Macro-F1 of 82.4\%; therefore, we use \mbox{DeepSeek-1.5B} as the default backbone in the remaining experiments unless otherwise specified.

\textbf{Module Configurations and Optimization.} In the Evidence-Enhanced Fusion Projector (MGE-EFP), the high-frequency excitation is driven by a differentiable Laplacian operator. The Relation-Aware Graph Network (R-AUGNN) is deliberately constrained to $L=2$ graph convolutional layers to capture multi-hop AU interactions while preventing the over-smoothing of node features. During the training phase, the Counterfactual Consistency Regularization (CCR) is activated with empirically determined balancing coefficients $\lambda_{ccr}=0.1$.

The entire framework is optimized using the AdamW optimizer with a weight decay of 1e-4. We apply a differential learning rate strategy to stabilize cross-modal alignment: 2e-4 for the visual and graph neural modules, and a more conservative 1e-4 for the LLM LoRA parameters. To obtain stable gradient estimates for noisy and imbalanced micro-expression samples, we use a batch size of 256. Given the complexity of aligning multi-granular visual features and topological structures with the LLM's semantic space, the model requires an extended training schedule for convergence. The framework achieves its best observed performance after 350 training epochs. All experiments are accelerated using a single NVIDIA H100 Tensor Core GPU, which provides the necessary memory bandwidth and computational efficiency for LLM-based inference.

\begin{table}[t]
  \centering
  \caption{Controlled computational profile for the LLM replacement study. The full AULLM++ keeps the LLM backbone frozen and updates only LoRA, projector, graph, and classifier parameters. The non-LLM Transformer head is trained from scratch and parameter-matched to the trainable part of AULLM++.}
  \label{tab:cost_profile}
  \footnotesize
  \setlength{\tabcolsep}{3.5pt}
  \renewcommand{\arraystretch}{1.12}
  \resizebox{\columnwidth}{!}{%
  \begin{tabular}{lccc}
    \hline\hline
    \textbf{Model} & \textbf{Total Params} & \textbf{Trainable Params} & \textbf{Inference Note} \\
    \hline
    AULLM++ (Ours)
    & $\sim$1.81B
    & $\sim$34.9M
    & Frozen 1.5B LLM; no CCR at test time \\
    Param.-matched Transformer
    & 35.4M
    & 35.4M
    & $\sim$77M FLOPs/sample; 1.0--1.2 ms/sample \\
    \hline\hline
  \end{tabular}%
  }
\end{table}

We further report a controlled computational profile in Table~\ref{tab:cost_profile}. AULLM++ is not designed to be the most lightweight micro-expression AU detector; its total model footprint is dominated by the frozen 1.5B LLM backbone. However, only LoRA adapters and task-specific visual, graph, projection, and classification modules are optimized, resulting in about 34.9M trainable parameters. The parameter-matched Transformer baseline has a similar number of trainable parameters but removes the pre-trained LLM entirely. Therefore, the ablation in Table~\ref{tab:ablation_big} controls for trainable model capacity, while Table~\ref{tab:cost_profile} makes the accuracy--robustness--cost trade-off explicit. CCR is applied only during training and introduces no additional inference-time computation.

\subsection{Within-Domain Evaluation}

\begin{table}[t]
  \centering
  \caption{Micro-expression AU detection performance comparison on the 4DME-Micro dataset under the LOSO protocol. F1-scores (\%) are reported. The best results are highlighted in \textbf{bold}. Avg. denotes Macro-F1.}
  \label{tab:4dme}
  \scriptsize
  \setlength{\tabcolsep}{3pt}
  \renewcommand{\arraystretch}{1.2}

  \resizebox{\columnwidth}{!}{%
  \begin{tabular}{lccccccccc}
    \hline\hline
    \textbf{Method} & \textbf{AU1} & \textbf{AU2} & \textbf{AU4} & \textbf{AU6} & \textbf{AU7} & \textbf{AU12} & \textbf{AU17} & \textbf{AU45} & \textbf{Avg.} \\
    \hline
    ResNet18~\cite{resnet}  & \textbf{57.6} & 55.4 & 45.9 & 50.1 & 50.8 & \textbf{56.9} & 52.0 & 52.5 & 52.6 \\
    SSSNet LED~\cite{LED}   & 46.9 & 50.7 & 52.9 & 46.3 & 42.8 & 47.6 & 49.0 & 72.9 & 51.1 \\
    \hline
    AULLM (Conf.)~\cite{aullm_conf}           & 53.7 & 58.8 & 51.2 & \textbf{50.5} & \textbf{51.8} & 48.2 & 49.0 & 77.0 & 55.0 \\
    \textbf{AULLM++ (Ours)} & 54.1 & \textbf{59.1} & \textbf{56.4} & 50.0 & 48.6 & 49.0 & \textbf{62.5} & \textbf{79.5} & \textbf{57.7} \\
    \hline\hline
  \end{tabular}%
  }
\end{table}

We first evaluate the within-domain detection performance of AULLM++ across the CASME II, SAMM, and 4DME-Micro datasets under the rigorous Leave-One-Subject-Out (LOSO) cross-validation protocol. The comparative results against representative methods are summarized in the corresponding tables. The baselines include traditional handcrafted features, pure CNN architectures, knowledge distillation paradigms, recent motion magnification approaches (e.g., SSSNet LED), and our preliminary conference framework, AULLM.

\vspace{1.5ex}
\noindent\textbf{Comparison with Conventional Paradigms.} As demonstrated in the SAMM evaluation (Table~\ref{tab:au_samm_transposed}), AULLM++ achieves a competitive Macro-F1 score of 62.6\%, outperforming the motion magnification baseline SSSNet LED by 8.4 percentage points. A similar trend is observed on the CASME II dataset, where our framework reaches 82.4\% Macro-F1. While recent motion magnification techniques attempt to address the low signal-to-noise ratio by explicitly amplifying temporal changes, they remain confined to the traditional feature pooling and pure data-driven classification paradigm. Consequently, these methods still suffer from the dilution of fine-grained high-frequency evidence during deep network propagation and have limited ability to incorporate AU-relation priors. In contrast, AULLM++ explicitly disentangles and preserves localized high-frequency textures via the MGE-EFP, compressing them into compact semantic tokens. This suggests that combining visual evidence construction with structured-token-conditioned inference improves the recognition of subtle micro-expression AUs.

\begin{table}[t]
\centering
\caption{AU label sets used for cross-dataset evaluation. `C', `S', and `D' denote CASME II, SAMM, and 4DME-Micro, respectively. For each dataset pair, both transfer directions use the same evaluated AU intersection. AUs outside the listed set are excluded from Macro-F1 computation and are not treated as negative labels.}
\label{tab:tab_dataset}
\footnotesize
\setlength{\tabcolsep}{3.5pt}
\renewcommand{\arraystretch}{1.12}
\begin{tabular}{p{0.20\columnwidth}p{0.60\columnwidth}c}
\toprule
Dataset pair & Evaluated AU set & \#AUs \\
\midrule
C $\leftrightarrow$ S
& AU2, AU4, AU7, AU12
& 4 \\
C $\leftrightarrow$ D
& AU1, AU2, AU4, AU7, AU12, AU17
& 6 \\
S $\leftrightarrow$ D
& AU2, AU4, AU7, AU12
& 4 \\
\bottomrule
\end{tabular}
\end{table}

\vspace{1.5ex}
\noindent\textbf{Advancement over the Conference Version.} The evaluation on the challenging and recently introduced 4DME-Micro dataset (Table~\ref{tab:4dme}) highlights the architectural improvements of AULLM++ over our preliminary conference version, AULLM. While AULLM established a strong baseline (55.0\% Macro-F1) by introducing LLMs into micro-expression analysis, it treated different action units as isolated classification targets and lacked a mechanism to model the underlying physiological dependencies between facial muscles.

By integrating the Relation-Aware Action Unit Graph Neural Network (R-AUGNN) for FACS anatomical priors with the MGE-EFP for high-frequency texture enhancement, AULLM++ captures these co-occurring or mutually inhibitory muscle dynamics more effectively. For instance, the synergistic activation of AU6 (cheek raiser) and AU12 (lip corner puller) is explicitly modeled as a structural instruction. Consequently, AULLM++ increases the overall Macro-F1 score to 57.7\% on 4DME-Micro, achieving improvements on most evaluated AUs, while several AU-level results remain challenging. This supports the usefulness of modeling physical priors for relation-aware AU prediction in complex micro-expression scenarios.

\subsection{Generalization Evaluation}
\begin{table}[t]
  \centering
  \caption{Cross-dataset generalization (Macro-F1, \%). `C', `S', and `D' denote the CASME II, SAMM, and 4DME-Micro datasets, respectively. The arrow ($\rightarrow$) indicates the train $\rightarrow$ test direction. The best results are highlighted in \textbf{bold}.}
  \label{tab:crossdomain_big}
  \scriptsize
  \setlength{\tabcolsep}{4pt}
  \renewcommand{\arraystretch}{1.2}

  \resizebox{\columnwidth}{!}{%
  \begin{tabular}{l ccccccc}
    \hline\hline
    \textbf{Method} & \textbf{C $\rightarrow$ S} & \textbf{C $\rightarrow$ D} & \textbf{S $\rightarrow$ C} & \textbf{S $\rightarrow$ D} & \textbf{D $\rightarrow$ C} & \textbf{D $\rightarrow$ S} & \textbf{Avg.} \\
    \hline
    ResNet18~\cite{resnet}  & 37.3 & 42.3 & 47.4 & 39.3 & 44.9 & 39.1 & 41.7 \\
    SSSNet LED~\cite{LED}   & 36.5 & 42.7 & 46.1 & 37.7 & 31.5 & 33.4 & 38.0 \\
    \hline
    AULLM (Conf.)\cite{aullm_conf}           & 49.9 & 44.1 & 52.8 & 48.3 & 47.2 & 41.6 & 47.3 \\
    \textbf{AULLM++ (Ours)} & \textbf{51.3} & \textbf{48.9} & \textbf{54.1} & \textbf{52.6} & \textbf{55.0} & \textbf{49.3} & \textbf{51.9} \\
    \hline\hline
  \end{tabular}%
  }
\end{table}

Micro-expression detection typically suffers from noticeable performance degradation when applied to unseen domains, a challenge widely recognized in recent domain generalization and contrastive learning studies~\cite{domainME, constrastiveme}. This vulnerability stems from the significant domain shifts introduced by heterogeneous elicitation environments, varying camera specifications, and diverse subject demographics (e.g., the high ethnic diversity in SAMM compared to the predominant Asian demographic in CASME II). To rigorously evaluate cross-domain robustness, we conduct comprehensive transfer experiments across six dataset pairs, where models are trained on the source dataset and directly evaluated on the target dataset without any fine-tuning. Since the AU label spaces are not identical across CASME II, SAMM, and 4DME-Micro, we compute cross-dataset Macro-F1 only on the AU intersection shared by each source-target pair. For each pair in Table~\ref{tab:tab_dataset}, the same evaluated AU set is used for both transfer directions, e.g., CASME II $\rightarrow$ SAMM and SAMM $\rightarrow$ CASME II are both evaluated on {AU2, AU4, AU7, AU12}. Non-overlapping or missing AU labels are excluded from both the reported metric and error counting, rather than being treated as negative labels.

As detailed in Table~\ref{tab:crossdomain_big}, under this challenging setting, conventional data-driven CNNs (ResNet18) and motion magnification methods (LED-SSSNet) experience considerable performance drops. For instance, in the 4DME $\rightarrow$ CASME II transfer task, the Macro-F1 of LED-SSSNet decreases to 31.5\%. Because these models often lack explicit physical constraints, they tend to overfit to the source domain's statistical biases---such as specific illumination conditions or sensor noise---rather than capturing the underlying physiological mechanisms of micro-expressions.

Compared with these baselines, AULLM++ achieves consistent improvements across all six transfer protocols. Notably, in the challenging 4DME $\rightarrow$ CASME II and 4DME $\rightarrow$ SAMM tasks, AULLM++ improves Macro-F1 by 7.8 and 7.7 percentage points, respectively, over the preliminary conference version (AULLM). This improved generalization capability primarily arises from the combined effect of FACS-guided AU-relation modeling and intervention-inspired consistency regularization. At the structural level, human facial anatomical structures and muscle dependencies defined by FACS remain relatively stable across different ethnicities and environments. By integrating the R-AUGNN, AULLM++ guides the structured inference process with these AU-relation priors, encouraging the model to focus on the topological correlations of muscle activations rather than superficial pixel-level discrepancies.

Complementing this structural prior, the Counterfactual Consistency Regularization (CCR) acts as an intervention-inspired regularizer during training. By applying targeted structural perturbations, CCR helps mitigate residual pseudo-correlations between domain-specific environmental noise and AU labels. The combination of R-AUGNN's structural guidance and CCR's perturbation-based regularization encourages the LLM-based inference module to shift its learning focus from pattern memorization to AU-relation-aware prediction, leading to more reliable cross-dataset micro-expression generalization.

\subsection{Ablation Studies}

\begin{table}[t]
  \centering
  \caption{Ablation study of AULLM++ modules (Macro-F1, \%). The results assess the contribution of each architectural component across all three benchmarks. The best results are highlighted in \textbf{bold}.}
  \label{tab:ablation_big}
  \footnotesize
  \setlength{\tabcolsep}{4pt}
  \renewcommand{\arraystretch}{1.15}

  \resizebox{\columnwidth}{!}{%
  \begin{tabular}{l ccc}
    \hline\hline
    \textbf{Variant} & \textbf{CASME II} & \textbf{SAMM} & \textbf{4DME} \\
    \hline
    \textbf{Full Framework (AULLM++)} & \textbf{82.4} & \textbf{62.6} & \textbf{57.7} \\
    \hline
    \multicolumn{4}{@{}l}{\textit{Structural AU modeling}} \\
    w/o R-AUGNN (no $\tau_{au}$) & 79.8 & 61.3 & 55.5 \\
    FACS graph $\rightarrow$ fully connected & 81.6 & 61.8 & 56.0 \\
    FACS graph $\rightarrow$ self-loops only & 80.2 & 60.5 & 54.0 \\
    \hline
    \multicolumn{4}{@{}l}{\textit{Evidence construction}} \\
    w/o EFP (linear proj.) & 78.9 & 59.6 & 51.3 \\
    $F_{\text{mid}}$ only  & 80.2 & 60.4 & 56.4 \\
    $F_{\text{high}}$ only & 79.7 & 60.5 & 53.6 \\
    \hline
    \multicolumn{4}{@{}l}{\textit{LLM and robustness}} \\
    w/o LLM (parameter-matched Transformer head)& 79.9 & 57.6 & 52.8 \\
    w/o LLM (MLP head) & 78.3 & 57.2 & 49.8 \\
    w/o CCR            & 81.4 & 60.9 & 55.2 \\
    \hline\hline
  \end{tabular}%
  }
\end{table}

To analyze the performance gains and validate the necessity of each proposed core module, we conduct ablation studies across the CASME II, SAMM, and 4DME-Micro datasets. As detailed in Table~\ref{tab:ablation_big}, we independently evaluate the contributions of the structural AU modeling, visual evidence construction, and token-conditioned inference components.

We first investigate the impact of structural AU modeling. Removing the Relation-Aware Action Unit Graph Neural Network (R-AUGNN) entirely leads to a noticeable performance drop; for instance, the Macro-F1 score on 4DME-Micro decreases from 57.7\% to 55.5\%. To further verify the necessity of physical priors, we replace the FACS-guided sparse topology with a fully connected graph and a self-loops-only graph. Both variants yield suboptimal results compared with the full model. This suggests that learning relation weights from small-sample datasets without anatomical constraints can easily lead to overfitting, whereas explicit anatomical priors help regularize the graph structure.

\begin{figure}[t]
\centering
\includegraphics[width=\columnwidth]{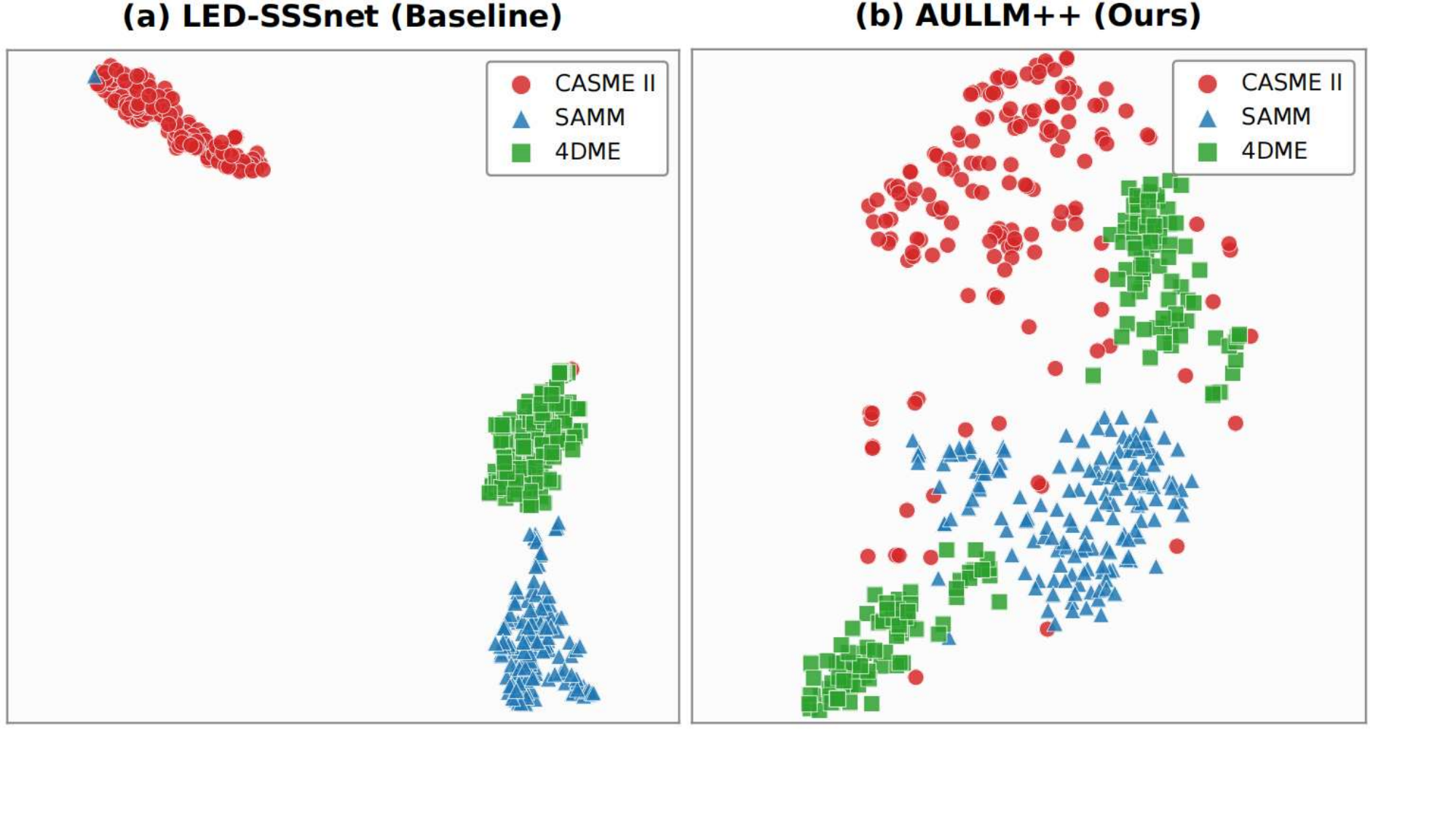}
\caption{t-SNE visualization of high-level features across CASME II, SAMM, and 4DME-Micro datasets. (a) The baseline LED-SSSNet exhibits clear domain shifts with isolated clusters. (b) AULLM++ shows better cross-domain feature alignment and feature entanglement.}
\label{fig:tsne}
\end{figure}
\begin{figure*}[t]
\centering
\includegraphics[width=\textwidth]{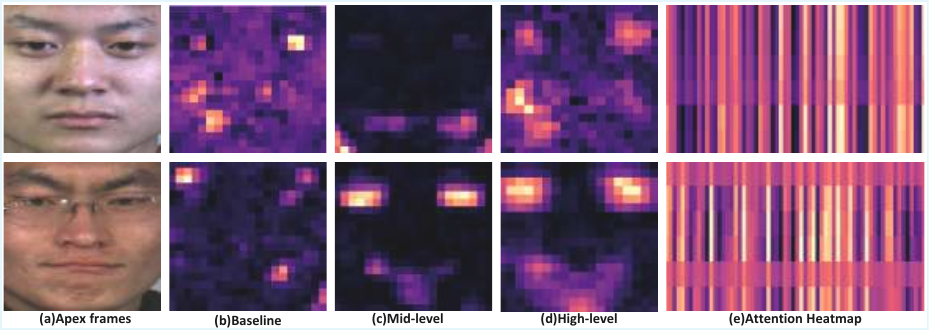}
\caption{Visualization of feature evolution.}
\label{fig:vizz}
\end{figure*}
Regarding visual evidence construction, the Multi-Granularity Evidence-Enhanced Fusion Projector (MGE-EFP) plays an important role in capturing subtle dynamics. When the attention-gated fusion is removed and replaced by a simple linear projection, the performance degrades across all benchmarks. Moreover, using only mid-level high-frequency features or only high-level global semantics fails to match the performance of the integrated module. This supports our hypothesis that accurate micro-expression detection benefits from the simultaneous extraction and adaptive fusion of localized edge deformations and overall facial context.

Finally, evaluating the LLM-based inference module reveals that replacing the DeepSeek-R1 LLM with a conventional MLP classification head results in the largest performance decline, such as dropping from 62.6\% to 57.2\% on the SAMM dataset. This emphasizes the practical advantage of mapping multimodal tokens into a structured token space for AU prediction rather than relying on standard multi-label regression. Additionally, removing the Counterfactual Consistency Regularization (CCR) leads to consistent declines across all datasets. This observation highlights its contribution to maintaining model stability and mitigating dataset-specific overfitting.

\subsection{Cross-Domain Feature Visualization}

To qualitatively illustrate the effectiveness of AULLM++ in mitigating cross-domain distribution shifts, we employ t-SNE to visualize the high-level feature spaces across the CASME II, SAMM, and 4DME-Micro datasets. As illustrated in Fig.~\ref{fig:tsne}(a), the feature distribution of the baseline method (LED-SSSNet) exhibits clear domain gaps, with isolated clusters formed according to dataset origin rather than physiological meaning. This indicates that the baseline model may overfit to dataset-specific environmental biases, such as lighting conditions, camera sensors, or ethnic backgrounds, which limits its ability to capture robust cross-domain micro-expression cues.

In contrast, Fig.~\ref{fig:tsne}(b) reveals that the feature distribution of AULLM++ shows a higher degree of domain overlap and entanglement. Driven by the FACS-guided AU-relation priors from R-AUGNN and the intervention-inspired regularization of CCR, our framework encourages the model to reduce dataset-specific environmental pseudo-correlations. Consequently, features from different datasets are projected into a closer and more generalized semantic space. While perfectly aligning distinct datasets into a single domain-agnostic manifold remains an open challenge, this visualization qualitatively suggests that AULLM++ learns better-aligned cross-domain micro-expression representations.

\subsection{Qualitative Analysis}

To illustrate the feature behavior and interpretability of AULLM++, we visualize the feature evolution process in Fig.~\ref{fig:vizz}. Compared with the diffuse and scattered activation patterns of the baseline 3D-CNN (Fig.~\ref{fig:vizz}b), which often drift toward irrelevant background noise or static facial contours, AULLM++ exhibits a clearer coarse-to-fine refinement trajectory. As illustrated, the mid-level features (Fig.~\ref{fig:vizz}c) capture transient high-frequency texture variations in micro-expressions, highlighting subtle edge deformations. The high-level features (Fig.~\ref{fig:vizz}d) then consolidate these fine-grained cues into coherent semantic regions. Driven by the R-AUGNN's structural instructions, this multi-granular integration enables the final attention map (Fig.~\ref{fig:vizz}e) to target anatomically significant muscles. For instance, the model localizes the \textit{zygomaticus major} for AU12 and isolates the relevant muscle groups for the complex AU6+12+17 combination. This suggests that our framework can disentangle subtle, co-occurring facial dynamics under physical guidance more effectively than purely data-driven visual baselines.

\section{Conclusion}
\label{sec:conclusion}

In this paper, we proposed AULLM++, a structured-token-conditioned framework for micro-expression action unit detection that integrates visual evidence with FACS-guided AU-relation priors. To address the challenges of low signal-to-noise ratios and severe domain shifts, we designed an architecture that combines visual evidence construction, structural AU modeling, and LLM-based token-conditioned prediction. The Multi-Granularity Evidence-Enhanced Fusion Projector extracts and fuses high-frequency local muscle variations with global facial context to form condensed visual tokens. Complementing this module, the Relation-Aware Action Unit Graph Neural Network embeds FACS-guided anatomical rules and provides structural AU-relation guidance for token-conditioned prediction. Furthermore, Counterfactual Consistency Regularization acts as an intervention-inspired training regularizer to reduce dataset-specific shortcut correlations. Extensive experiments on three benchmarks demonstrate that AULLM++ achieves competitive performance in both within-domain evaluations and challenging cross-dataset transfer tasks.

Despite these advances, complete domain generalization in micro-expression analysis remains an open challenge. In future work, we aim to develop a specialized micro-expression AU foundation model with stronger and more interpretable AU-relation modeling capabilities. By encouraging the model to provide more transparent evidence attribution for AU activations, we seek to further enhance the reliability, interpretability, and generalization of affective computing systems under complex real-world conditions.

\bibliographystyle{IEEEtran}
\bibliography{ref}

\begin{IEEEbiography}[{\includegraphics[width=1in,height=1.25in,clip,keepaspectratio]{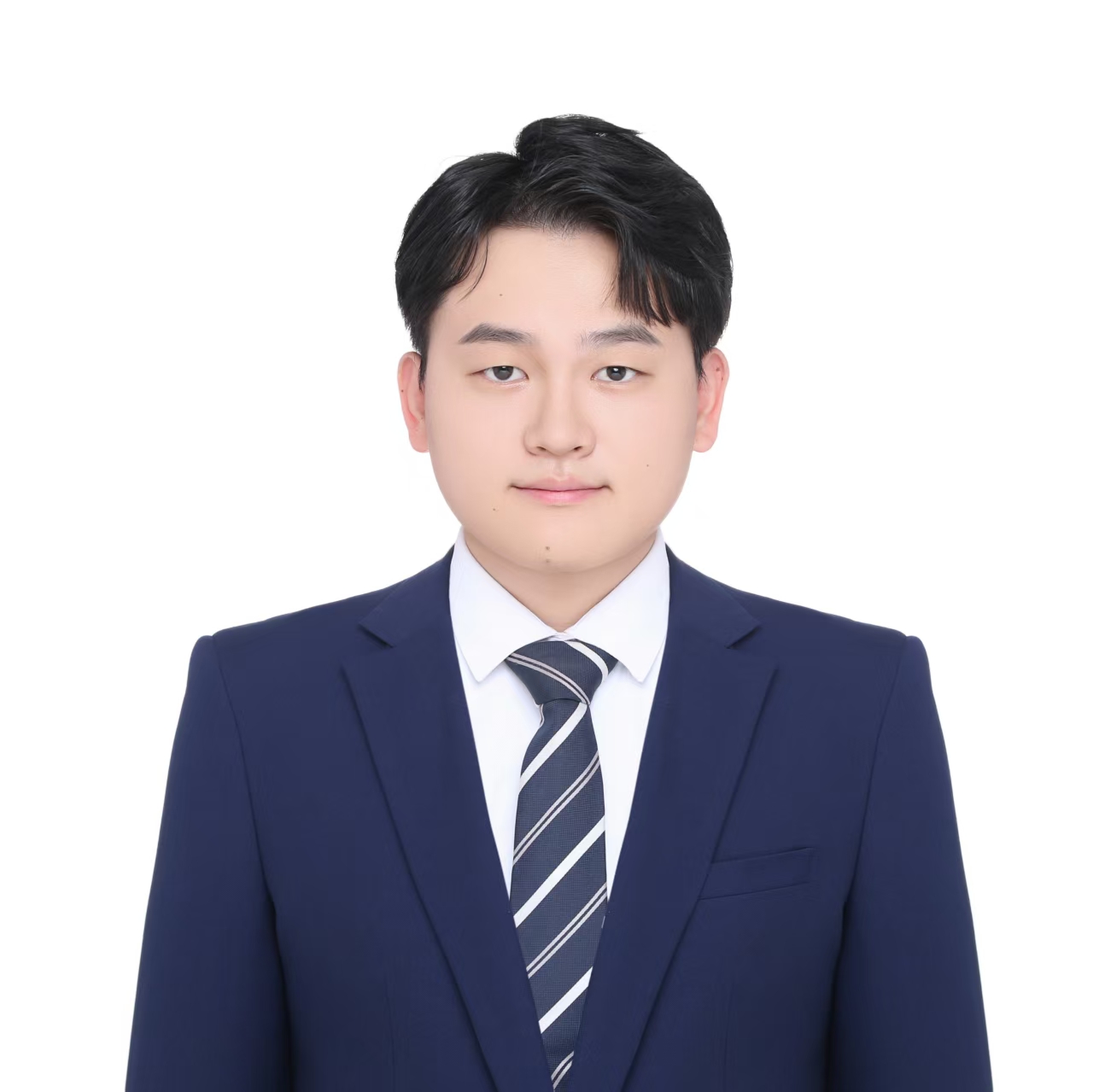}}]{Zhishu Liu}
received the B.Eng. degree from the College of Artificial Intelligence, Jilin University, Changchun, China. He is currently an M.Sc. student in computer science at City University of Hong Kong (Dongguan), Dongguan, China, and a visiting student at Great Bay University, Dongguan, China. His current research interests include micro-expression action unit detection, micro-expression recognition, and multimodal large models for affective computing.

\end{IEEEbiography}

\begin{IEEEbiography}[{\includegraphics[width=1in,height=1.25in,clip,keepaspectratio]{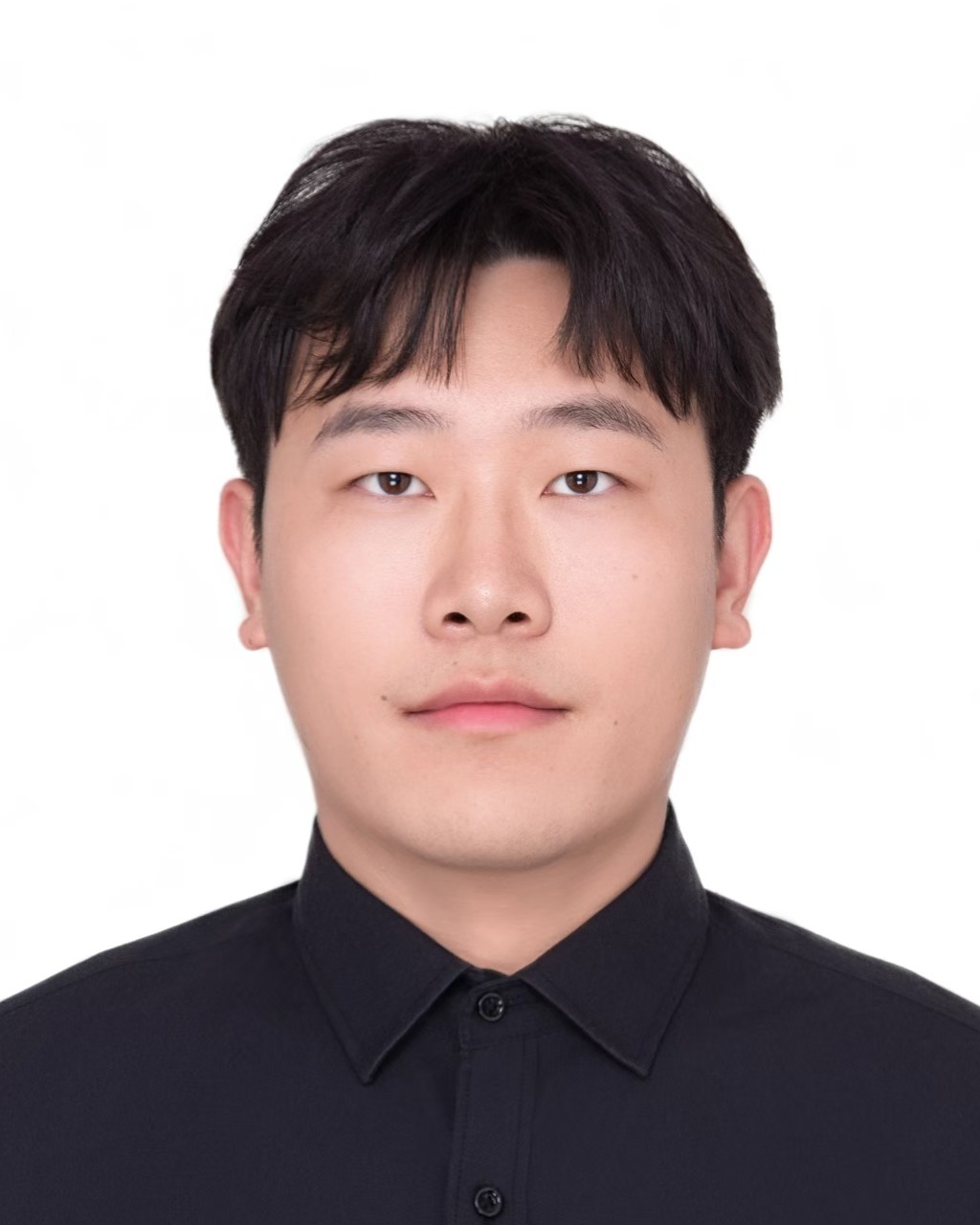}}]{Kaishen Yuan}
received the B.S. and M.S. degrees from Tianjin University, China. He is currently pursuing the Ph.D. degree in artificial intelligence at The Hong Kong University of Science and Technology (Guangzhou). His research interests include affective computing, facial analysis, and multimodal large language model reasoning. He has published several papers in conferences and journals, including ECCV, ICLR, CVPR, ACM MM, and IEEE TIP.
\end{IEEEbiography}

\begin{IEEEbiography}[{\includegraphics[width=1in,height=1.25in,clip,keepaspectratio]{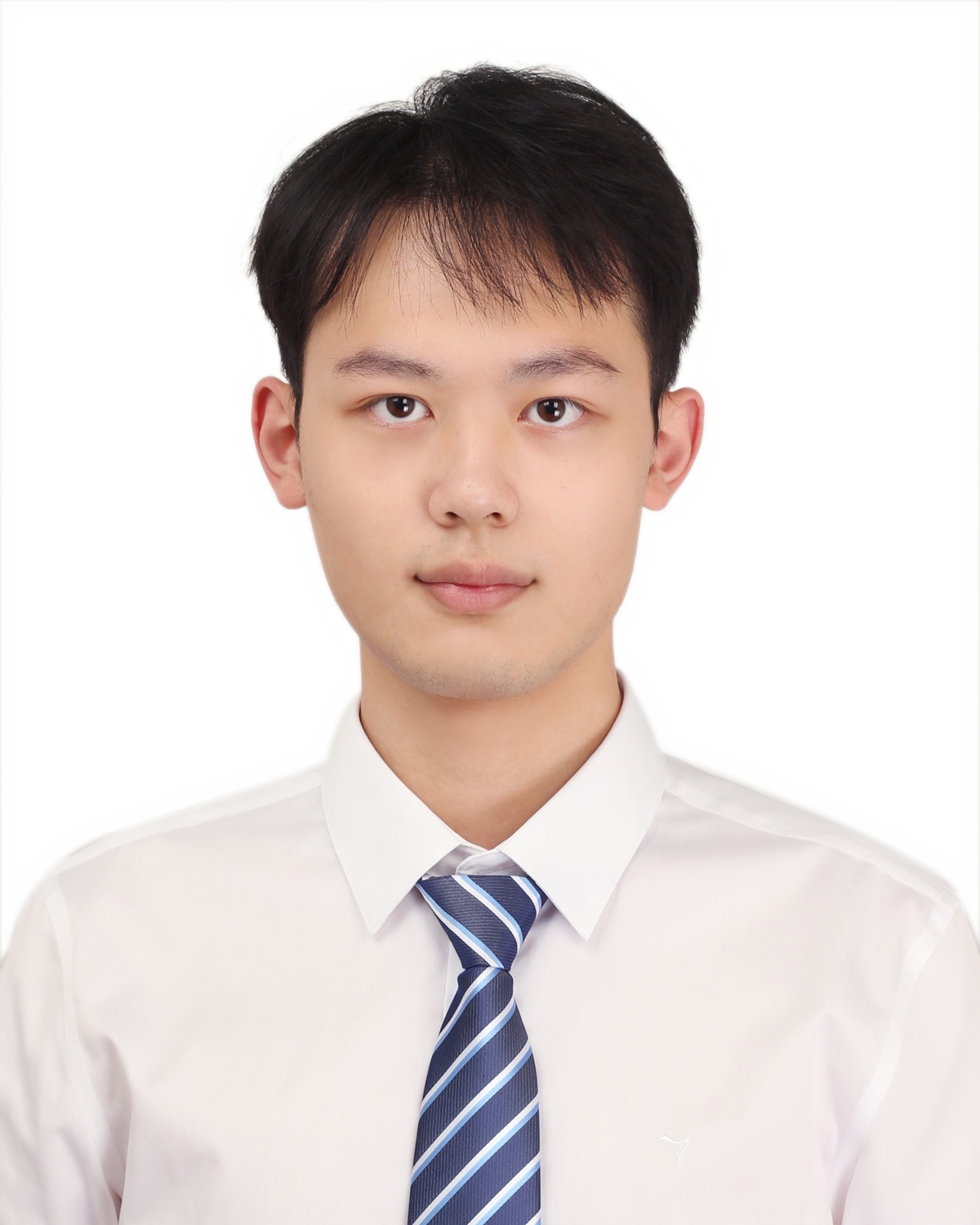}}]{Bo Zhao}
is currently pursuing the master's degree in artificial intelligence and robotics at The Chinese University of Hong Kong, Shenzhen. He is also a visiting student at Great Bay University. His research interests include remote photoplethysmography, multimodal large language models, and agentic memory. His research has been published in venues such as ICLR and CVPR.
\end{IEEEbiography}

\begin{IEEEbiography}[{\includegraphics[width=1in,height=1.25in,clip,keepaspectratio]{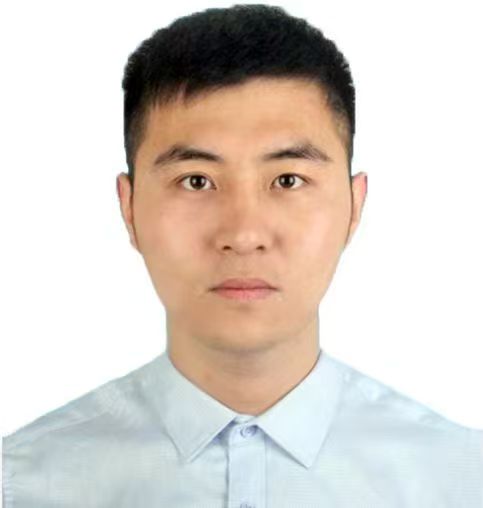}}]{Hui Ma}
received the M.S. degree in science from the Macau University of Science and Technology (MUST), Macau, China, where he is currently working toward the Ph.D. degree. His research interests include deep learning, computer vision, domain generalization, and multimodal face anti-spoofing.

\end{IEEEbiography}

\begin{IEEEbiography}[{\includegraphics[width=1in,height=1.25in,clip,keepaspectratio]{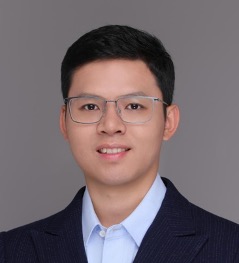}}]{Zitong Yu}
(Senior Member, IEEE) received the Ph.D. degree in computer science and engineering from the University of Oulu, Finland, in 2022. He is currently a tenured Associate Professor at Great Bay University, China. He was a postdoctoral researcher at ROSE Lab, Nanyang Technological University, and a visiting scholar at TVG, University of Oxford, from July to November 2021. His research interests focus on subtle visual computing. He received the IAPR Best Student Paper Award and the IEEE Finland Section Best Student Conference Paper Award.
\end{IEEEbiography}
\end{document}